\documentclass[letterpaper, 10 pt, conference]{ieeeconf}  

\usepackage[T1]{fontenc}
\usepackage[utf8]{inputenc}
\usepackage{textcomp}
\usepackage{amsmath}
\usepackage{amssymb}
\usepackage{microtype}
\usepackage{newtxtext,newtxmath}  
\usepackage{graphicx}
\usepackage{booktabs}
\usepackage{float}
\usepackage{subfigure}
\usepackage{caption}
\usepackage{multirow}

\DeclareMathOperator*{\argmin}{arg\,min}
\DeclareMathOperator*{\argmax}{arg\,max}

\newcommand{\insertfig}{\includegraphics[width=\linewidth,height=100pt]{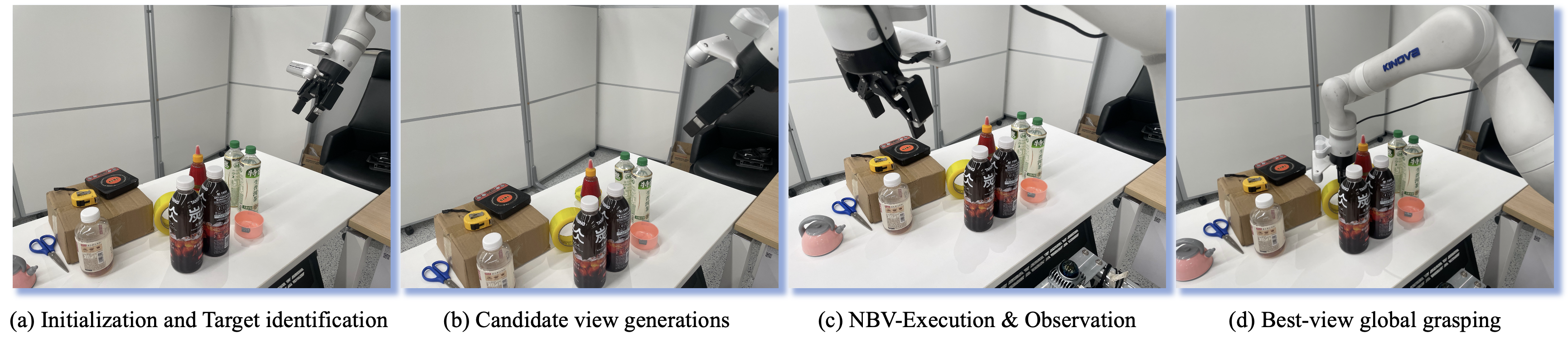}\captionof*{figure}{\textbf{GraspView} integrates a perception scene robotic grasping in cluttered environments. The active perception scoring module selects the best viewpoint to construct a global perception scene under occlusions. Integrated with the VLM-guided best-view grasping, it enables perception grasping through a unified perception–decision–execution pipeline.}}

\IEEEoverridecommandlockouts                              

\title{\LARGE \bf
GraspView: Active Perception Scoring and Best-View Optimization for Robotic Grasping in Cluttered Environments
}

\author{
Shenglin Wang$^{1,*}$, 
Mingtong Dai$^{1,2,4,*}$, 
Jingxuan Su$^{3}$, 
Lingbo Liu$^{1,\dagger}$, 
Chunjie Chen$^{2}$, 
Xinyu Wu$^{2}$, 
Liang Lin$^{1,5,6}$
\thanks{$^{1}$Peng Cheng Laboratory, Shenzhen, China. $^{2}$Shenzhen Institute of Advanced Technology, Chinese Academy of Sciences, Shenzhen, China. $^{3}$Shenzhen Graduate School, Peking University, Shenzhen, China. $^{4}$University of Chinese Academy of Sciences, Beijing, China. $^{5}$School of Computer Science and Engineering, Sun Yat-sen University, Guangzhou, China. $^{6}$X-Era AI Lab, China.}%
\thanks{*Equal contribution. $^{\dagger}$Corresponding author: Lingbo Liu ({\tt\small liulb@pcl.ac.cn})}%
}

\makeatletter
\apptocmd{\@maketitle}{\centering\insertfig}{}{}
\makeatother

\begin{document}

\maketitle

\thispagestyle{empty}
\pagestyle{empty}

\begin{abstract}

Robotic grasping is a fundamental capability for autonomous manipulation, yet remains highly challenging in cluttered environments where occlusion, poor perception quality, and inconsistent 3D reconstructions often lead to unstable or failed grasps.
Conventional pipelines have widely relied on RGB-D cameras to provide geometric information, which fail on transparent or glossy objects and degrade at close range.
We present GraspView, an RGB-only robotic grasping pipeline that achieves accurate manipulation in cluttered environments without depth sensors. Our framework integrates three key components: (i) global perception scene reconstruction, which provides locally consistent, up-to-scale geometry from a single RGB view and fuses multi-view projections into a coherent global 3D scene; (ii) a render-and-score active perception strategy, which dynamically selects next-best-views to reveal occluded regions; and (iii) an online metric alignment module that calibrates VGGT predictions against robot kinematics to ensure physical scale consistency. Building on these tailor-designed modules, GraspView performs best-view global grasping, fusing multi-view reconstructions and leveraging GraspNet for robust execution.
Experiments on diverse tabletop objects demonstrate that GraspView significantly outperforms both RGB–D and single-view RGB baselines, especially under heavy occlusion, near-field sensing, and with transparent objects. These results highlight GraspView as a practical and versatile alternative to RGB–D pipelines, enabling reliable grasping in unstructured real-world environments.

\end{abstract}

\section{INTRODUCTION}


Robotic grasping constitutes a fundamental capability in robotics, and the ability to execute reliable grasps in cluttered environments is essential for enabling robots to interact effectively with diverse and unstructured objects~\cite{fang2020graspnet, bircher2016receding,luo2025precise}. However, in cluttered environments, robotic grasping faces significant challenges, including limited scene reconstruction, metric scale inconsistency, and inadequate cross-view consistency~\cite{ASGrasp,huang2024copa}. These issues significantly reduce the effectiveness of multi-view fusion and grasp planning, thereby hindering precise and reliable grasping in real-world scenarios. Addressing these challenges is essential for improving robotic performance in cluttered and occluded environments. Therefore, this research is highly relevant and necessary, as it aims to overcome these limitations by introducing novel solutions that enhance scene reconstruction, active perception, and grasp planning strategies.
Traditional methods for grasping scene reconstruction typically rely on RGB-D cameras to obtain 3D point clouds, which suffer from depth inaccuracies and high costs~\cite{schwarz2018rgb}. Currently, mainstream multi-view reconstruction methods do not depend on depth cameras~\cite{an2024rgbmanip}. However, these methods often require a large number of initial views and struggle with the issue of metric scale consistency between point clouds and the robot's coordinate frame. The issues of inconsistent metric scale and poor cross-view consistency substantially undermine the effectiveness of multi-view fusion and grasp planning. Recently introduced, VGGT is a state-of-the-art model that supports both single and multi-image inputs, enabling joint prediction of point clouds, depth maps, and camera poses~\cite{vggt}. Furthermore, VGGT’s depth map predictions are up-to-scale, and when combined with camera extrinsics they provide a strong basis for building point clouds and downstream grasping. In practice, to achieve physically correct scale in the robot base frame, we enforce online metric alignment during execution. Despite these advancements, reconstructing a global 3D scene from limited initial views remains a significant challenge.

Post-reconstruction robot active perception and precise grasping remain major challenges in robotic grasping. Existing active perception methods typically rely on fixed strategies, where the viewpoint is selected based on predetermined criteria, and the grasping action follows directly from this fixed selection~\cite{bousmalis2018using}. Consequently, the accuracy of grasping is entirely dependent on the performance of the fixed strategy. These methods fail to incorporate dynamic viewpoint selection or adapt based on real-time visual perception, limiting the system’s ability to adjust to the changing scene and choose the optimal viewpoint~\cite{jin2024robotgpt,ji2025robobrain}. As a result, the lack of flexibility in viewpoint selection undermines the potential for robust and precise grasping.


To address the aforementioned challenges, we propose \textbf{GraspView}, a global perception robotic grasping framework in cluttered environments. The framework consists of three core modules: a global perception scene construction module, an active perception scoring module, and a best-view global grasping module. The scene construction module initializes a local 3D map from a single image and generates multi-view projections, which are selected by a vision-language model (VLM) and reconstructed into a global 3D scene using VGGT. The active perception scoring module leverages VLM to evaluate projected images, selecting the top-ranked viewpoints as next-best-view (NBV) candidates. The robot then moves to the NBV for observation, where online metric consistency is enforced between the real-world scene and reconstructed point cloud before executing grasp actions. Finally, the best-view global grasping module employs the VLM again to select the globally optimal viewpoint based on the reconstructed global perception scene, enabling precise and robust grasping.

In summary, the contributions of this paper are as follows:
\begin{itemize}
  \item \textbf{Global 3D scene reconstruction.} We build a global 3D scene from a single-view initialization by selecting informative multi-view projections via a vision–language model (VLM) and fusing them with VGGT, yielding a coherent point-cloud representation for downstream grasping.
  \item \textbf{Scoring-based active perception.} We adopt a render-and-score strategy that off-screen renders candidate viewpoints and uses a VLM to score visibility/grasp utility, selecting the highest-scoring next-best view (NBV) to reduce occlusion efficiently.
  \item \textbf{Online metric alignment.} We recover a global scale factor by aligning VGGT-predicted relative motions with robot kinematics, enforcing metric consistency of the reconstructed geometry for reliable planning and execution.
\end{itemize}

\section{RELATED WORK}
\subsection{Monocular 3D Reconstruction for Grasping}
Robotic grasping pipelines traditionally rely on RGB--D sensors to obtain metric scene geometry, but depth sensing can be brittle on glossy or transparent surfaces and adds nontrivial hardware cost\cite{ASGrasp}. Multi-view RGB pipelines based on structure-from-motion and multi-view stereo (e.g., COLMAP\cite{schoenberger2016sfm,schoenberger2016mvs}) combined with neural radiance fields or 3D Gaussian Splatting\cite{nerf,3dgs} have recently been explored for grasping, including open-vocabulary variants via semantic distillation\cite{Decomposingnerf,featuresplatting,feature3dgs} and grasp-centric extensions\cite{gaussiangrasper,graspsplats}. These approaches can avoid depth sensors but often require many views and long optimization cycles, and may complicate metric alignment.

Single-view depth and geometry predictors offer a complementary path. Forward models such as DPT and DUSt3R\cite{dpt,dust3r} estimate dense depth or point maps directly from RGB, but scale ambiguity and cross-view inconsistency can hinder multi-view fusion and downstream planning. VGGT\cite{vggt} departs from this trend by jointly predicting camera extrinsics and dense geometry (depth/point maps) from one or a few RGB frames, albeit up-to-scale, substantially reducing the friction of building RGB-only, point-cloud–driven grasping systems. We leverage this capability to collapse reconstruction and pose estimation into a single module that interfaces cleanly with grasp planners, while recovering metric consistency online during active perception.

\subsection{Active Perception for Grasping}
Active perception selects viewpoints that reduce occlusion and improve task success. Classical NBV methods design hand-crafted utilities (e.g., coverage, information gain) to guide view planning for reconstruction\cite{pan2024viewsneededreconstructunknown,activegs} or grasping\cite{MVPicking,breyer2022closedloopnextbestviewplanningtargetdriven,ACE-NBV,VISO-Grasp}. While effective, fixed scoring rules can be brittle across categories and clutter. In contrast, we adopt a render-and-score strategy: we render candidate viewpoints off-screen from the current colored point cloud and use a vision-language model to score each view for expected visibility and grasp utility, then execute only the top-rated pose. This removes the need for bespoke heuristics, can incorporate semantic cues, and selects a single informative move before acquiring a real image.

\section{METHOD}

\begin{figure*}[t]
  \centering
  \includegraphics[width=1\textwidth]{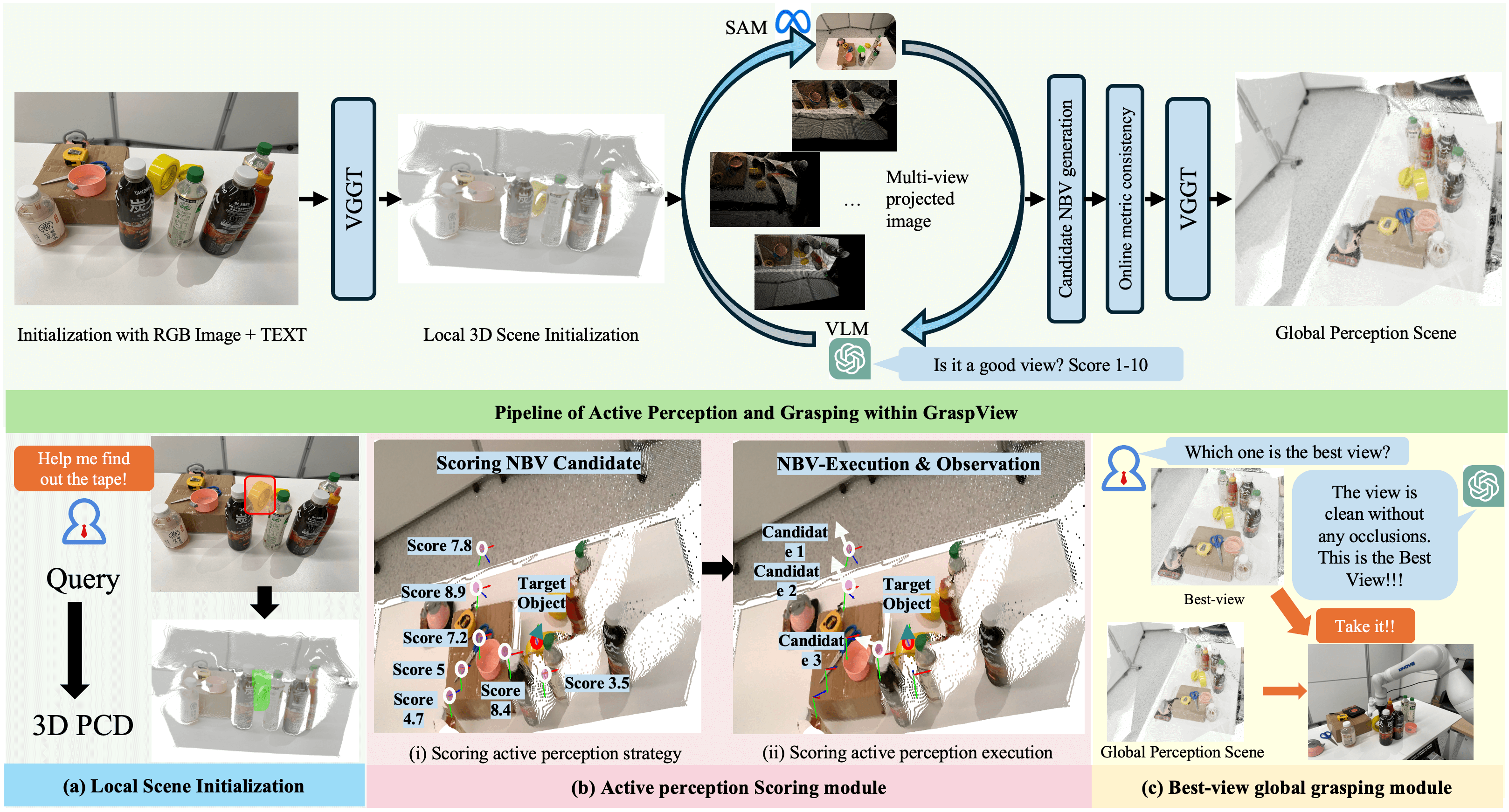}
  \caption{Overview of Our GraspView framework. It comprises three modules: (a) global perception scene construction module, (b) active perception scoring module, (c) best-view global grasping module. }
  \label{fig:pipeline}
\end{figure*}

\subsection{Overview}
Figure~\ref{fig:pipeline} presents the overall pipeline of \textbf{GraspView}. The framework first performs local 3D scene initialization from an RGB image and language query to obtain a metrically consistent point cloud. Next, the active perception module employs SAM and a vision-language model (VLM) to score and select next-best-view (NBV) candidates, ensuring informative observations under occlusions. Finally, the best-view global grasping module utilizes the reconstructed global scene to identify the most suitable viewpoint and execute reliable grasp poses. This unified pipeline enables robust grasping in cluttered environments while supporting open-vocabulary queries.


\subsection{Local Scene Initialization Module}

\textbf{Text-to-3D Local Scene Initialization.}  
Vision–language models (VLMs) provide strong image–language alignment and interactive reasoning capabilities, enabling a robot to infer human intent from natural-language prompts (e.g., “grasp the red bottle”). However, while VLMs excel at semantic understanding in 2D, they inherently lack metric spatial reasoning in 3D. To bridge this gap, we integrate VLM-based grounding with geometric reconstruction, thereby mapping language-conditioned intent into a localized 3D representation.  

Given an image $I_t$ and a query $q$, the VLM produces spatial prompts  

\begin{equation}
\pi_{\mathrm{VLM}} = g_{\mathrm{VLM}}(I_t, q),
\end{equation}

which encode approximate 2D support of the referred object (e.g., bounding boxes, point seeds, or coarse regions). These prompts are passed to a promptable segmentation backbone (e.g., SAM2), yielding a high-fidelity binary mask  

\begin{equation}
M_t = \mathrm{Seg}(I_t;\,\pi_{\mathrm{VLM}},\,\theta).
\end{equation}

Concurrently, we run VGGT on the initial view $I_0$ to obtain a dense point map $\mathcal{P}_0^W$ and extrinsic $\mathbf{T}_{C_0W}$. Transforming into the robot base frame via $\mathbf{T}_{BW}$ produces $\mathcal{P}_0^B$, establishing a metrically consistent local 3D scene. The segmentation mask $M_t$ is resampled to the VGGT grid and back-projected with known intrinsics $\mathbf{K}$ and extrinsics $\mathbf{T}_{C_0B}$, filtering the initialized cloud into an object-centric subset  

\begin{equation}
\mathcal{P}_{\mathrm{obj}}^{(t)} = \Big\{\, \mathbf{x}_B \in \mathcal{P}_0^{B}\;\Big|\; \Pi\!\left(\mathbf{K}\,\mathbf{T}_{C_0B}\,\mathbf{x}_B\right) \in \Omega(M_t) \,\Big\}.
\end{equation}

with centroid  

\begin{equation*}
\mathbf{c} 
= \dfrac{1}{\big\lvert\mathcal{P}_{\mathrm{obj}}^{(t)}\big\rvert}\,\sum_{\mathbf{x}_B\in\mathcal{P}_{\mathrm{obj}}^{(t)}} \mathbf{x}_B.
\end{equation*}

This pipeline establishes a text-to-3D mapping: the VLM grounds semantic intent in 2D, SAM refines the object mask, and VGGT projects it into a metrically consistent 3D workspace. The resulting local scene reconstruction provides both (i) object-specific point clouds for occlusion-aware reasoning and (ii) a partial 3D map of the environment to guide the sampling of next-best-view (NBV) candidates.  

\textbf{NBV Candidate Generation.} We instantiate next-best-view (NBV) candidates by sampling camera poses around the target centroid $\mathbf{c}\in\mathbb{R}^3$ in the base frame using a \emph{conical} spherical-cap sampler around the current line of sight.

\paragraph{Conical sampling} Let the current camera center be $\mathbf{p}_0\!=\!(\mathbf{T}_{BC_0})_{t}$ and define the viewing axis from object to camera and the default radius
\begin{equation}
\mathbf{a} = \frac{\mathbf{p}_0 - \mathbf{c}}{\lVert \mathbf{p}_0 - \mathbf{c} \rVert},\quad r = \lVert \mathbf{p}_0 - \mathbf{c} \rVert.
\end{equation}
Construct an orthonormal basis $\{\mathbf{u}_0,\mathbf{u}_1,\mathbf{u}_2\}$ with $\mathbf{u}_0\!\parallel\!\mathbf{a}$ and $\mathbf{u}_1,\mathbf{u}_2$ spanning the plane orthogonal to $\mathbf{u}_0$. Given a half-angle $\alpha_{\max}$, discretize
\begin{align}
\alpha_0 &= 0,\; \alpha_i = \tfrac{i}{n_{\alpha}}\,\alpha_{\max}\; (i{=}1..n_{\alpha}),\\
\beta_j &= \tfrac{2\pi j}{n_{\beta}}\; (j{=}0..n_{\beta}{-}1).
\end{align}
For each pair $(\alpha_i,\beta_j)$, define the unit direction on the spherical cap and the candidate position as
\begin{align}
\mathbf{d}_{i,j} &= \cos\alpha_i\,\mathbf{u}_0 + \sin\alpha_i\,(\cos\beta_j\,\mathbf{u}_1 + \sin\beta_j\,\mathbf{u}_2),\\
\mathbf{p}_{i,j} &= \mathbf{c} + r\,\mathbf{d}_{i,j}.
\end{align}
The axial sample $\alpha_0$ uses a single azimuth ($j{=}0$). The total number of samples is $1 + n_{\alpha}n_{\beta}$.

\paragraph{Orientation and roll stabilization} For any sampled position $\mathbf{p}$, we use a look-at rotation pointing the camera $+\mathbf{z}$ axis to the target. Let the reference up direction be the initial camera's up in $\mathcal{F}_B$,
\begin{equation}
\mathbf{u}_{\mathrm{ref}} = \mathbf{R}_{BC_0}\,\mathbf{e}_y,\quad \mathbf{e}_y{=}[0,1,0]^\top.
\end{equation}
Define forward, right, and true-up unit vectors as
\begin{equation}
\mathbf{f}(\mathbf{p}) = \frac{\mathbf{c} - \mathbf{p}}{\lVert \mathbf{c} - \mathbf{p} \rVert},\;\;
\mathbf{r}(\mathbf{p}) = \frac{\mathbf{u}_{\mathrm{ref}}\times\mathbf{f}(\mathbf{p})}{\lVert \mathbf{u}_{\mathrm{ref}}\times\mathbf{f}(\mathbf{p}) \rVert},\;\;
\hat{\mathbf{u}}(\mathbf{p}) = \mathbf{f}(\mathbf{p})\times\mathbf{r}(\mathbf{p}).
\end{equation}
The camera-to-base rotation and pose are
\begin{align}
\mathbf{R}_{BC}(\mathbf{p}) &= \big[\,\mathbf{r}(\mathbf{p})\;\; \hat{\mathbf{u}}(\mathbf{p})\;\; \mathbf{f}(\mathbf{p})\,\big],\\
\mathbf{T}_{BC}(\mathbf{p}) &=
\begin{bmatrix}
\mathbf{R}_{BC}(\mathbf{p}) & \mathbf{p} \\
\mathbf{0}^\top & 1
\end{bmatrix}.
\end{align}
This stabilizes roll across candidates and maintains a consistent wrist orientation, matching the implementation choice that uses the initial camera to define the up-reference. The conical sampler biases candidates towards minimal motion around the current vantage point, reducing kinematic issues and move time.

\paragraph{Render-and-Score NBV selection} Given each feasible candidate $\mathbf{T}_{BC}(\mathbf{p})$, we synthesize an off-screen rendering $\hat{I}_{\mathbf{p}}$ of the current colored object cloud using intrinsics $\mathbf{K}$ and a pinhole model (with near/far planes set by workspace limits). We color points by back-projecting their appearance from the latest RGB using the known extrinsics, and splat them with Z-buffering to obtain a clean 2D view. A vision--language model (VLM) then scores each rendering for task utility. We prompt for visibility and grasp suitability and combine them into a scalar score
\begin{equation}
S(\mathbf{p}) = w_{v}\,s_{\mathrm{vis}}(\hat{I}_{\mathbf{p}}) + w_{g}\,s_{\mathrm{grasp}}(\hat{I}_{\mathbf{p}}) - w_{o}\,s_{\mathrm{occl}}(\hat{I}_{\mathbf{p}}),
\end{equation}
where $w_{v},w_{g},w_{o}\!\ge\!0$ and the individual scores are normalized to $[0,1]$. We retain only kinematically feasible candidates (IK + joint/velocity limits) and select the NBV as $\mathbf{p}^* = \argmax_{\mathbf{p}} S(\mathbf{p})$. The robot executes only $\mathbf{p}^*$ to acquire a real image $I_1$.

\subsection{Active Perception Scoring Module}

\textbf{Scoring Active Perception Strategy.} Active perception begins by sampling a set of kinematically feasible next-best-view (NBV) candidates around the target object. Each candidate viewpoint is rendered off-screen from the current colored point cloud, and a vision–language model (VLM) evaluates the rendered images with respect to visibility, grasp suitability, and occlusion. The NBV is selected as the highest-scoring candidate
\begin{equation}
v^{\ast}_{\mathrm{NBV}} = \argmax_{v_i} S(v_i),
\end{equation}
where $S(v_i)$ is the VLM-derived score balancing visibility, grasp utility, and occlusion penalties. The robot executes only $v^{\ast}_{\mathrm{NBV}}$ to capture a real RGB image, which, together with the initial wrist view $I_0$, is processed by VGGT for two-view fusion.

Since VGGT expresses geometry in the coordinate system of the first frame, the NBV image $I_{\mathrm{BV}}$ is used as the reference, yielding per-view extrinsics and dense point maps. Two synchronized representations are maintained: (i) a base-frame scene cloud $\mathcal{P}^B$ for visualization, workspace clipping, and collision checks, and (ii) a BV-camera-frame object cloud $\mathcal{P}^{BV}$ for GraspNet inference. Segmentation masks from both views are fused to extract the target object cloud $\mathcal{P}_{\mathrm{obj}}^{\ast}$, which is then refined by statistical filtering and clustering.

\begin{figure}[!tbh]
    \centering
    \includegraphics[width=\linewidth]{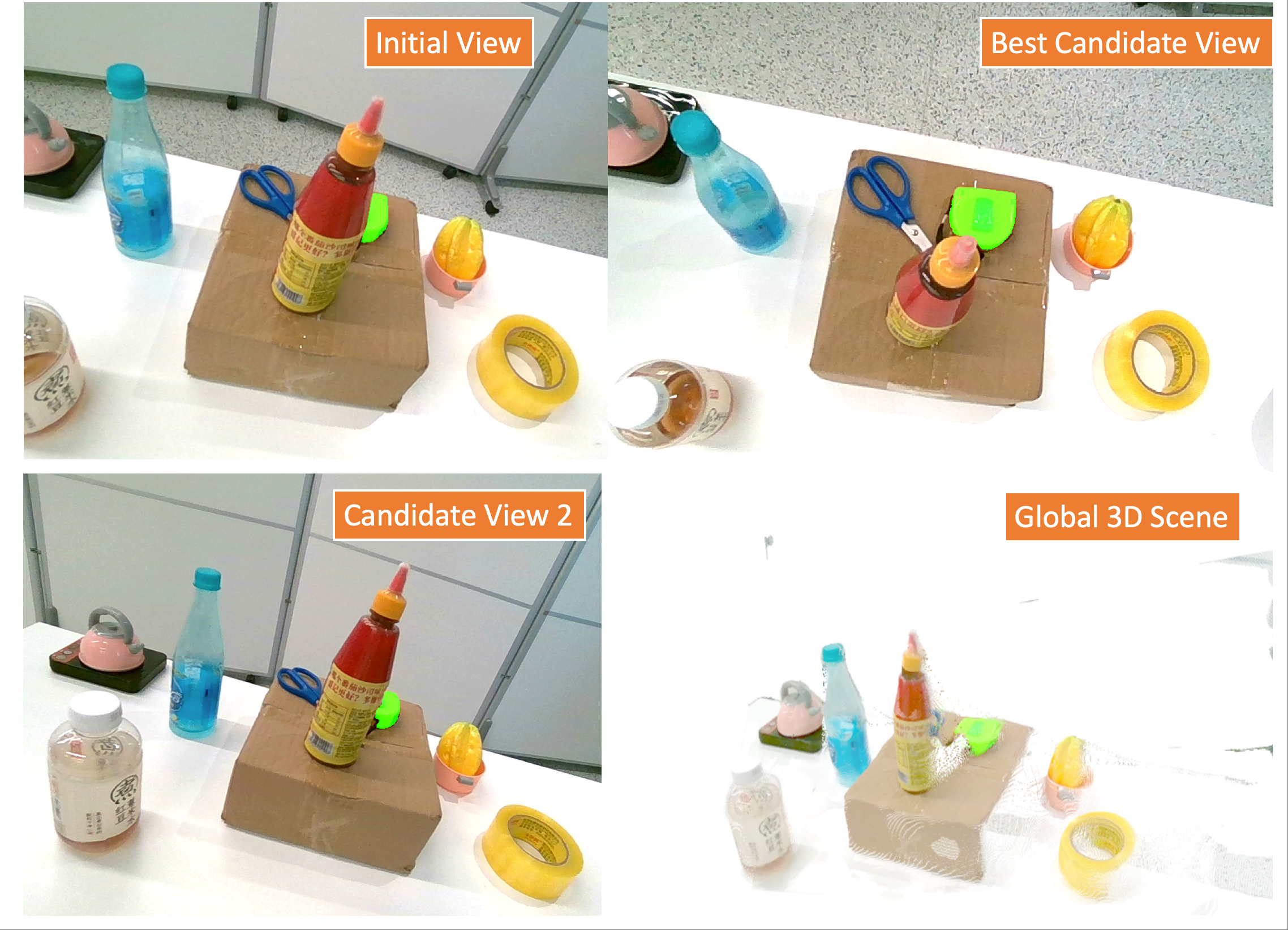}
    \caption{Multi-view candidate observations of tabletop objects and the fused occlusion-aware global point cloud reconstruction via VGGT. The integration of complementary views alleviates single-view occlusion, yielding a consistent 3D representation that supports robust object localization and grasp planning.}
    \label{fig:mv}
\end{figure}

\textbf{Scoring Active Perception Execution.} Let $I_{\mathrm{BV}}$ denote the executed Best-View (BV) image selected by the NBV policy, and let $I_0$ be the initial wrist image. Because VGGT expresses geometry in the coordinate system of the \emph{first} input frame, we place $I_{\mathrm{BV}}$ first in the multi-view call and use the ordering $\{I_{\mathrm{BV}}, I_0\}$. This anchors the reconstruction on the BV camera frame, matching our downstream grasp planning and execution, which also start from the BV pose. Fig.~\ref{fig:mv} shows the executions of active perception where the camera captures the target ruler from different views and captured images are fused into VGGT to get the global 3D scene.

The VGGT call returns per-view camera extrinsics and dense point maps. Empirically, GraspNet expects camera-frame point clouds -- using base-frame clouds led to unreasonable predictions. Thus, We maintain two synchronized representations: (i) a scene cloud in the robot base frame $\mathcal{F}_B$ for visualization, workspace clipping, and collision checking, obtained via the hand--eye calibrated camera-to-base transform $\mathbf{T}_{BC}^{\mathrm{BV}}$; and (ii) an object cloud in the BV camera frame $\mathcal{F}_{C}^{\mathrm{BV}}$ for GraspNet inference.

For object extraction, we resize both views' segmentation masks to the VGGT grid and keep points consistent with either mask (logical OR), giving precedence to BV when masks disagree near boundaries. We clip points to workspace bounds (e.g., within 1.0 m of the base origin), apply statistical and radius outlier removal, and run DBSCAN clustering, retaining the largest connected component as the final target cloud $\mathcal{P}_{\mathrm{obj}}^{\ast}$. The cloud provided to GraspNet is expressed in $\mathcal{F}_{C}^{\mathrm{BV}}$. If a base-frame cloud $\mathbf{x}_B$ is available, its relation to BV camera coordinates is
\begin{equation}
\mathbf{x}_{C}^{\mathrm{BV}} = (\mathbf{R}_{BC}^{\mathrm{BV}})^{\top}\,(\mathbf{x}_B - \mathbf{t}_{BC}^{\mathrm{BV}}),\quad \mathbf{T}_{BC}^{\mathrm{BV}}{=}[\,\mathbf{R}_{BC}^{\mathrm{BV}}\;|\;\mathbf{t}_{BC}^{\mathrm{BV}}\,].
\end{equation}

\begin{figure*}[t]
    \centering
    \includegraphics[width=\linewidth]{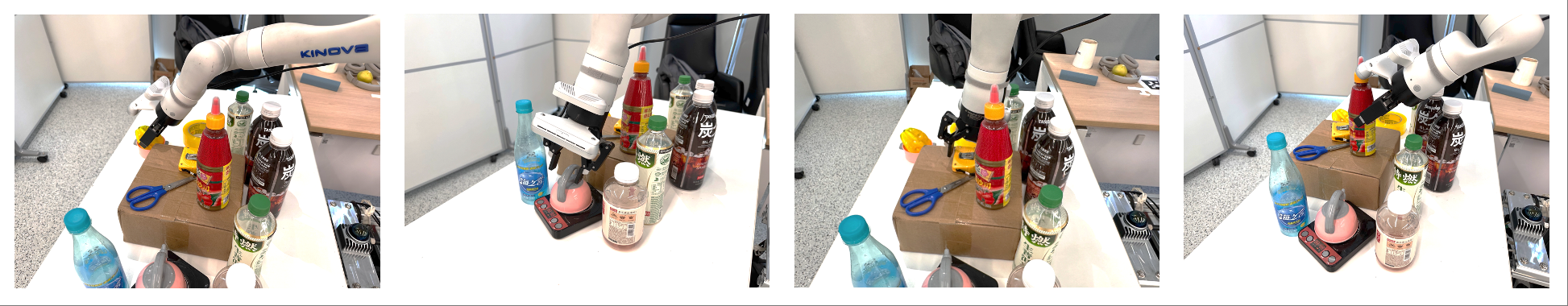}
    \caption{Grasping execution in cluttered tabletop scenes. The robotic arm interacts with diverse objects, including bottles, a kettle, a ruler, and fruits, under partial occlusions. These examples demonstrate the ability of the proposed system to perform reliable grasping in dense and visually complex environments.}
    \label{fig:ov}
\end{figure*}

\textbf{Online Metric Consistency via Active Perception.}
\label{sec:online_metric}
Although VGGT jointly estimates point maps and camera extrinsics, its predictions are inherently up-to-scale, resulting in discrepancies with the physical metric. To ensure consistency, we adopt an online scale recovery procedure inspired by recent joint calibration approaches~\cite{zhi2024unifying}. During active perception, we maintain two synchronized pose sequences: (i) the ground-truth camera poses $\mathbf{T}^{B}_{C,k}$ provided by robot kinematics in the base frame $\mathcal{F}_{B}$, and (ii) the VGGT-predicted relative extrinsics $\mathbf{T}^{C}_{W,k}$ expressed in the coordinate system of the first input frame, here anchored at the BV image.

For each executed NBV image $I_k$, we define the relative camera translations from both modalities as
\begin{equation}
\begin{split}
\Delta \mathbf{t}^{\mathrm{rob}}_{k} 
  &= \big(\,(\mathbf{T}^{B}_{C,0})^{-1}\,\mathbf{T}^{B}_{C,k}\,\big)_{t}, \\
\Delta \mathbf{t}^{\mathrm{vggt}}_{k} 
  &= \big(\,(\mathbf{T}^{C}_{W,0})^{-1}\,\mathbf{T}^{C}_{W,k}\,\big)_{t}.
\end{split}
\end{equation}
where $(\cdot)_{t}$ extracts the translation component. Since VGGT recovers only relative geometry up to an unknown scale factor $\lambda > 0$, we estimate $\lambda$ by solving a least-squares alignment:
\begin{equation}
\lambda^{\star} = \argmin_{\lambda>0} \sum_{k=1}^{M}
\big\lVert
\Delta \mathbf{t}^{\mathrm{rob}}_{k} - \lambda\,\Delta \mathbf{t}^{\mathrm{vggt}}_{k}
\big\rVert_2^{2}.
\end{equation}
This admits a closed-form solution:
\begin{equation}
\lambda^{\star} =
\frac{\sum_{k=1}^{M} \langle \Delta \mathbf{t}^{\mathrm{vggt}}_{k},\, \Delta \mathbf{t}^{\mathrm{rob}}_{k} \rangle}
{\sum_{k=1}^{M}\lVert\Delta \mathbf{t}^{\mathrm{vggt}}_{k}\rVert_2^{2}}.
\end{equation}

After recovering $\lambda^{\star}$, we scale the VGGT point maps and transform them into the BV camera frame by
\begin{equation}
\mathbf{p}^{C_{\mathrm{BV}}} =
\mathbf{R}_{cw}\,\big(\lambda^{\star}\,\mathbf{p}^{W}\big) + \mathbf{t}_{cw},
\end{equation}
where $[\,\mathbf{R}_{cw}\;|\;\mathbf{t}_{cw}\,] = \mathbf{T}^{C}_{W,\mathrm{BV}}$ denotes the BV camera extrinsic. The resulting metrically consistent point cloud $\mathcal{P}_{\mathrm{obj}}^{\ast}$ is used for GraspNet inference.

Notably, this scale recovery requires only a small number of NBV-selected views (typically one to three), enabling online alignment of VGGT reconstructions to real-world scale without extensive offline calibration. As a result, downstream grasp planning operates on physically accurate geometry, thereby enhancing execution reliability in cluttered and occluded scenes.

\subsection{Best-View Global Grasping Module}
Building upon the active perception stage, all executed views are fused via VGGT into a global 3D reconstruction $\mathcal{P}^{\mathrm{global}}$, capturing both the object and its surrounding environment. The previously traversed candidate viewpoints are re-scored by the VLM to identify an occlusion-free grasp view $v^{\ast}$:
\begin{equation}
v^{\ast} = \argmax_{v_i} S'(v_i),
\end{equation}
where $S'(v_i)$ emphasizes complete visibility of the target object. The target object cloud $\mathcal{P}_{\mathrm{obj}}^{\ast}$ is extracted in the grasp-view camera frame $\mathcal{F}_C^{v^{\ast}}$ and fed into GraspNet to generate grasp candidates and scores.

A predicted grasp pose $\mathbf{T}_{CE}$ in the grasp-view camera frame is mapped to the robot base frame by
\begin{equation}
\mathbf{T}_{BE} = \mathbf{T}_{BC}^{v^{\ast}}\,\mathbf{T}_{CE},
\end{equation}
where $\mathbf{T}_{BC}^{v^{\ast}}$ denotes the camera-to-base transformation of the grasp-view. Candidate grasp poses are refined into executable end-effector motions by (i) aligning GraspNet’s canonical gripper frame with the physical gripper through discrete symmetry rotations, and selecting the orientation closest to the current wrist configuration, and (ii) imposing a gravity-aligned top-down preference by aligning the approach axis with world $-\mathbf{z}$.

Execution is \emph{BV-anchored}: beginning at the grasp-view wrist pose, the robot plans a Cartesian pre-grasp at offset $d$ along the approach axis, checks inverse kinematics (IK) feasibility and collisions against $\mathcal{P}^{\mathrm{global}}$, and performs a straight-line approach, grasp closure, and vertical lift. If the selected grasp is infeasible, the system falls back to lower-ranked grasps or re-enters the NBV stage.

By combining global scene reconstruction with occlusion-free viewpoint selection, as shown in Fig.~\ref{fig:ov} the proposed module surpasses traditional RGB–D based 6Dof GraspNet pipelines~\cite{fang2020graspnet}, which operate on partial point clouds and often fail under occlusion. Our approach captures complete object geometry and environmental context, enabling robust grasping even in cluttered and heavily occluded scenes.

\begin{figure}[t]
    \centering
    \includegraphics[width=\linewidth]{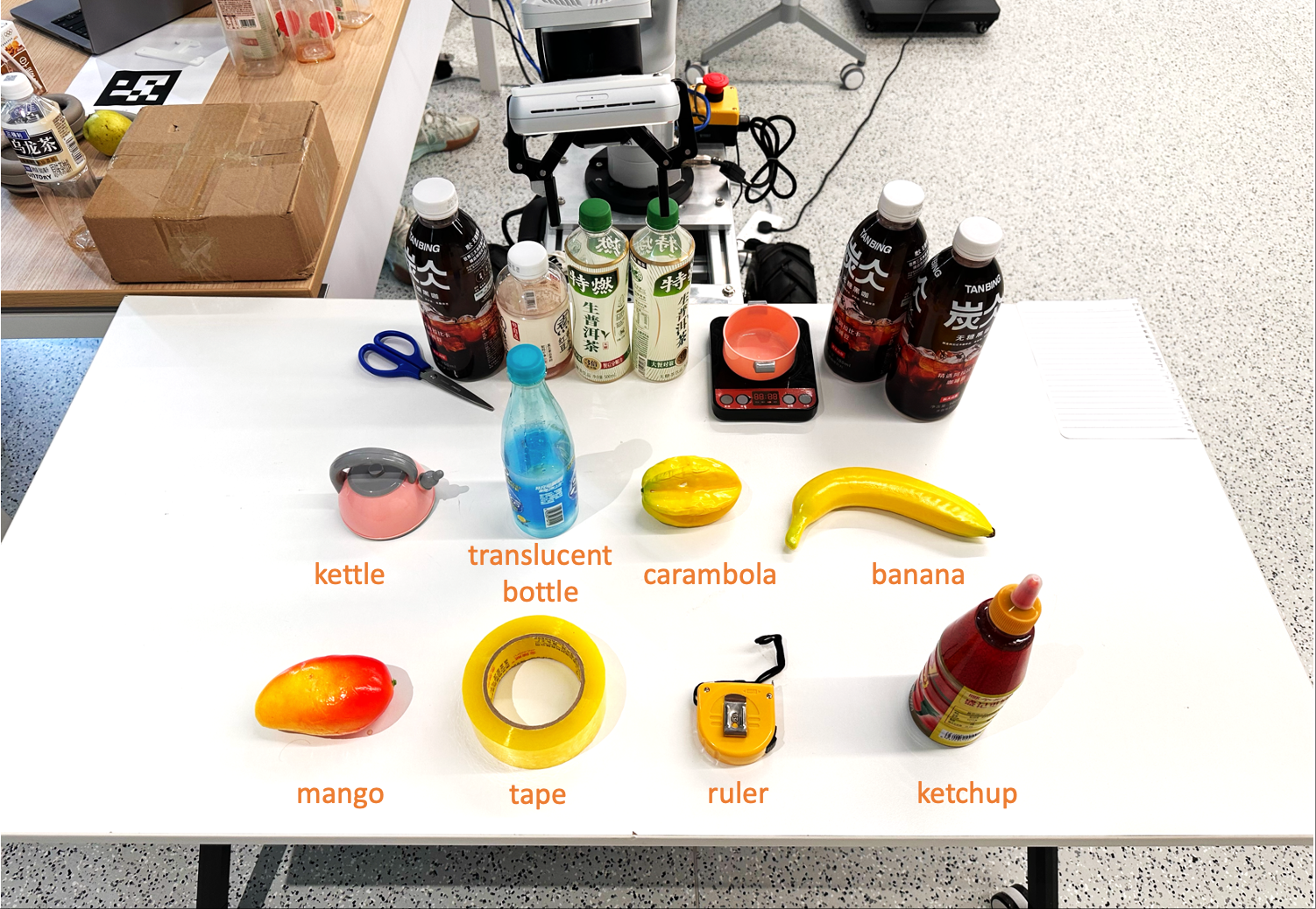}
    \caption{Experimental setup of grasping objects. 
    The labeled items on the table (\textit{kettle, translucent bottle, carambola, banana, mango, tape, ruler, ketchup}) are defined as the grasping targets. 
    The remaining items are placed as distractors or obstacles to increase the difficulty of perception and grasp planning.}
    \label{fig:tgobj}
\end{figure}

\section{EXPERIMENTS}

\begin{table*}[t]
  \centering
  \caption{Overall performance per object without/with occlusions. Columns report success rate (SR, \%) for each object and the mean across objects.}\resizebox{0.8\linewidth}{!}{
  \label{tab:overall_wocc}

\begin{tabular}{c|cccccccc|c}
  \toprule
  \multirow{2}*{Method} & \multicolumn{8}{c|}{w/o occ.} & \multirow{2}*{Avg SR} \\
  \cmidrule(lr){2-9}
   & kettle & bottle & carambola & banana & mango & tape & ruler & ketchup & \\
  \midrule
  RGB-D (SV)      & 60 & 50 & 65 & 65 & 60 & 55 & 60 & 70 & 60.625 \\
  GraspView (SV)  & 70 & 80 & 70 & 80 & 85 & 75 & 65 & 90 & 76.875 \\
  GraspView (MV)  & 75 & 90 & 70 & 80 & 90 & 75 & 70 & 90 & 80 \\
  \bottomrule
\end{tabular}}

\vspace{0.5em}
\resizebox{0.8\linewidth}{!}{
\begin{tabular}{c|cccccccc|c}
  \toprule
  \multirow{2}*{Method} & \multicolumn{8}{c|}{w/ occ.} & \multirow{2}*{Avg SR} \\
  \cmidrule(lr){2-9}
   & kettle & bottle & carambola & banana & mango & tape & ruler & ketchup & \\
  \midrule
  RGB-D (SV)      & 30 & 10 & 30 & 20 & 30 & 15 & 15 & 35 & 23.125 \\
  GraspView (SV)  & 40 & 30 & 45 & 30 & 35 & 30 & 20 & 35 & 33.125 \\
  GraspView (MV)  & 75 & 85 & 60 & 75 & 80 & 75 & 70 & 80 & 75 \\
  \bottomrule
\end{tabular}}

\end{table*}

\subsection{System Setup}
Our experimental platform is based on a Kinova Gen3 manipulator with a wrist-mounted Intel RealSense D455 camera. To emphasize the monocular perception pipeline, the camera is operated strictly in RGB mode without depth sensing. Factory-provided intrinsics are used, and the rigid camera-to-end-effector transformation is calibrated via standard hand–eye calibration, ensuring metric consistency across frames. For perception, VGGT serves as the 3D geometry backbone, inferring dense point maps and camera poses directly from RGB inputs. Object-level segmentation is included as a general module—supporting both interactive (e.g., SAM/SAM2) and language-guided (e.g., DINO-X) approaches—but since segmentation is not the focus of our method, we do not elaborate on its details. Grasp proposals are generated using GraspNet, and the selected grasp is executed on the robot.

\subsection{Experiment Settings}

We design tabletop grasping experiments with multiple daily objects with different sizes and shapes
as the grasping targets, while additional items on the table serve as obstacles that partially occlude the targets 
(Fig.~\ref{fig:tgobj}). 

We evaluate the system under three experimental protocols:

\begin{enumerate}
    \item \textbf{Comparison across object shape and size.}  
    We compare the grasp success rate of different objects under two occlusion conditions: 
    \textit{w/o occlusion} and \textit{w/ occlusion}. 
    The baselines include: an RGB-D single-view pipeline, 
    GraspView without active perception (static single-view fusion), 
    and GraspView with active perception (NBV-based multi-view fusion).

    \item \textbf{Effect of NBV number and occlusion level.}  
    We vary the number of executed NBVs and evaluate performance under different occlusion severities: 
    \textit{easy} (target mostly visible), 
    \textit{medium} (roughly half of the target visible), 
    and \textit{hard} (only a small portion of the target visible). 
    This setting highlights the contribution of active perception in progressively uncovering occluded regions.

    \item \textbf{Impact of camera distance.}  
    To analyze the effect of viewing distance, we conduct experiments at 30cm, 50cm, 
    and 70cm between the camera and the workspace. 
    The grasp success rates of RGB-D and GraspView are reported to investigate robustness to scale and depth variation.
\end{enumerate}

\begin{figure*}
    \centering
    \includegraphics[width=\linewidth]{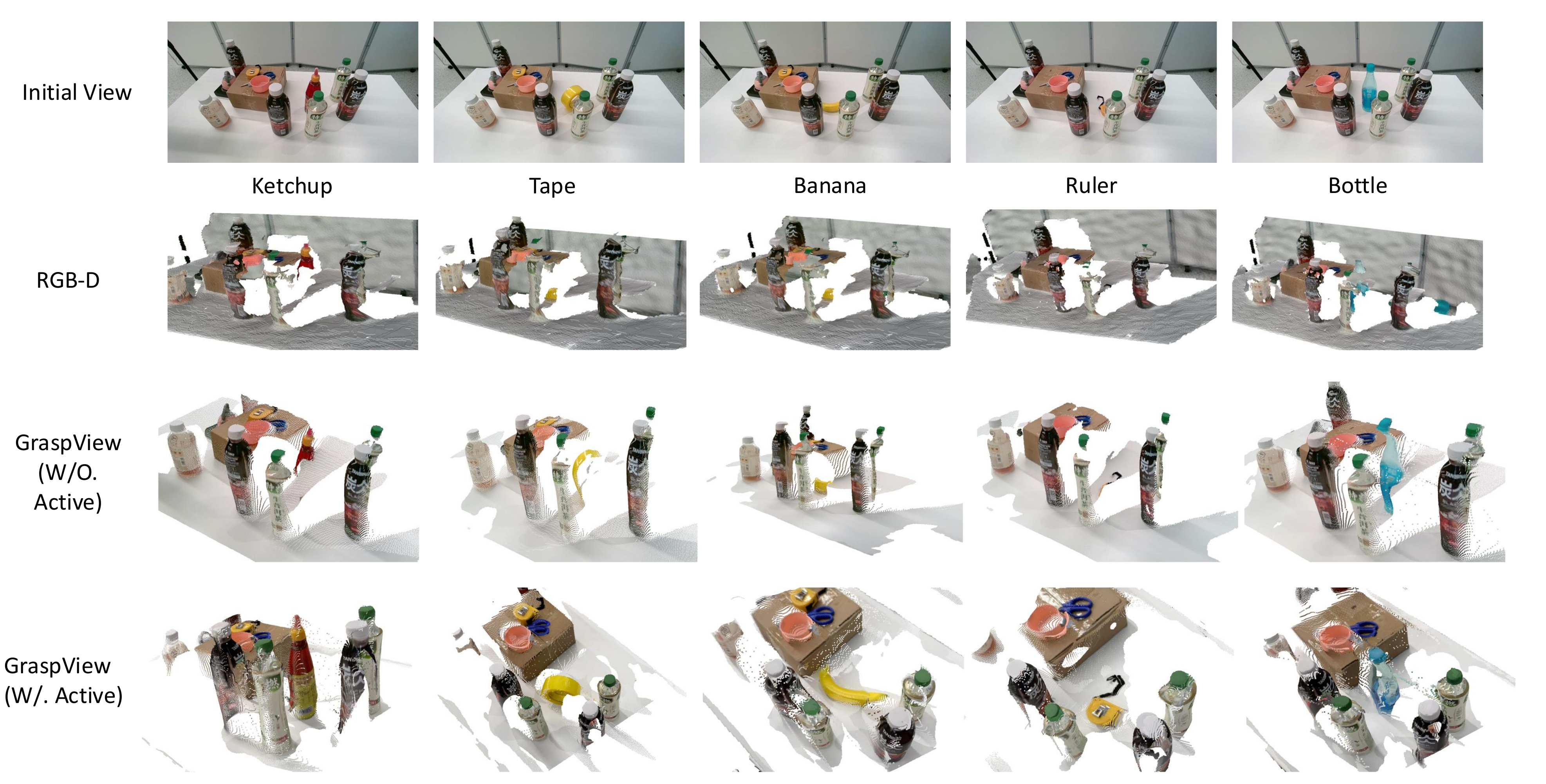}
    \caption{Qualitative results of tabletop grasping under occlusion. 
        The \textit{Initial View} illustrates the scene as observed by the wrist-mounted camera. 
        \textit{RGB-D} reconstruction suffers from incomplete or noisy geometry, particularly on transparent or occluded objects. 
        \textit{GraspView (w/o Active)} denotes the single-view variant, where limited visibility restricts accurate 3D reasoning. 
        In contrast, \textit{GraspView (w/ Active)} leverages NBV-based active perception to capture previously occluded regions (e.g., ruler), 
        leading to a more complete and metrically consistent 3D reconstruction for grasp planning.}
    \label{fig:visexp}
\end{figure*}

\subsection{Results}
\textbf{Overall performance across sensing setups and occlusions.} Table~\ref{tab:overall_wocc} reports per-object success rate (SR, \%) on eight tabletop objects (kettle, bottle, carambola, banana, mango, tape, ruler, ketchup) under two settings: w/o occlusion and w/ occlusion. Without occlusion, GraspView (MV) achieves the best average SR across all objects. The single-view GraspView (SV) attains 76.875\% on average and remains competitive on easier objects. In contrast, the RGB--D (SV) baseline averages 60.625\%. Under occlusion, both single-view configurations deteriorate substantially, whereas GraspView (MV) maintains markedly higher SR (75\% average). The visualization of parts of targets grasping is shown in Fig.~\ref{fig:visexp}.

\begin{table}[t]
  \centering
  \caption{Success rate (Avg SR, \%) across occlusion difficulty levels for different numbers of executed NBV explorations.}\resizebox{0.8\linewidth}{!}{
  \label{tab:nbv_occlusion}
  \begin{tabular}{l|l|ccc}
    \toprule
    Index & NBVs  & Easy & Medium & Hard \\
    \midrule
    A& 0 & 80 & 65 & 20 \\
    B& 1 & 90 & 80 & 55 \\
    C& 2 & 95 & 85 & 70 \\
    D& 3 & 95 & 90 & 70 \\
    \bottomrule
  \end{tabular}}
\end{table}

\textbf{Effect of occlusion difficulty and NBV budget.}
Table~\ref{tab:nbv_occlusion} illustrates how increasing the number of NBV explorations systematically improves grasp success under occlusion. 
In our evaluation, we categorize occlusion difficulty into three levels. 
\emph{Easy} cases involve minimal or negligible blockage, where the target object is almost fully visible. 
\emph{Medium} cases present partial occlusion, with certain important regions of the object obscured by surrounding clutter. 
\emph{Hard} cases correspond to severe occlusion, where the object is nearly hidden from the initial view and only small portions remain visible.
While the single-view baseline struggles when parts of the object are hidden, introducing even a single NBV already brings a large boost by revealing critical geometry. A second NBV provides further, though smaller, gains, particularly in the hard-occlusion setting. Beyond two explorations, the benefit largely saturates, suggesting that most useful information can be captured within one to two targeted viewpoint changes.
This highlights a practical trade-off: modest exploration achieves near-optimal robustness, whereas excessive moves add latency with little return.

\textbf{Sensitivity to working distance and modality.}
Table~\ref{tab:distance} compares RGB--D perception from a RealSense D455 with our RGB-only GraspView under different working distances. The depth-based pipeline suffers severely at close range, where the sensor cannot reliably reconstruct geometry, and its performance is further degraded by material properties such as transparency. In contrast, GraspView remains stable across all distances, delivering consistent success rates by relying on monocular geometry prediction rather than active depth sensing. At larger distances both methods converge, but only the RGB-only pipeline provides robustness across the full operating envelope, making it a more versatile option for unstructured environments.

\begin{table}[t]
  \centering
  \caption{Success rate (SR, \%) under different working distances for RGB--D (RealSense D455) vs. our RGB-only GraspView.}\resizebox{0.9\linewidth}{!}{
  \label{tab:distance}
  \begin{tabular}{l|ccc}
    \toprule
    Perception & 30\,cm & 50\,cm & 70\,cm \\
    \midrule
    RealSense (RGB-D) & 10 & 40 & 85 \\
    GraspView (VGGT)  & 80 & 85 & 85 \\
    \bottomrule
  \end{tabular}}
\end{table}

\textbf{Impact of world-coordinate alignment.}
We analyze how aligning VGGT geometry to the robot base affects pose accuracy. Table~\ref{tab:alignment} compares a no-alignment baseline (file hand--eye with a fixed global scale) against our online metric consistency alignment (Sec.~\ref{sec:online_metric}): a short image--motion sequence is used to recover a global scale and enforce consistency between predicted and observed camera poses. We report the recovered global scale ($\lambda$) and median frame-wise pose errors between predicted and observed base-to-camera poses (rotation in degrees, translation in meters). The alignment step reduces pose error while recovering a realistic scale, which yields more accurate grasp execution in practice.

\begin{table}[t]
    \centering
\caption{Effect of world-coordinate alignment on pose accuracy. $\lambda$ is the recovered global scale via the online metric consistency step. Rot and Trans are median frame-wise errors between predicted and observed base-to-camera poses (degrees, meters). For the no-alignment baseline, $\lambda$ is fixed to 1.00 (no scale estimation).}\resizebox{0.8\linewidth}{!}{
    \label{tab:alignment}
    \setlength{\tabcolsep}{6pt}
    \small
    \begin{tabular}{lccc}
        \toprule
        Setting & $\lambda$ & rot (deg) & trans (m) \\
        \midrule
        w/o align & 1.00 & 3.46 & 0.13 \\
        w align & 0.85 & 0.72 & 0.11 \\
        \bottomrule
    \end{tabular}}
\end{table}

\section{CONCLUSION}
We introduced \textbf{GraspView}, an RGB-only robotic grasping framework that integrates global 3D scene reconstruction, scoring-based active perception, and online metric alignment to enable robust grasping without depth sensors. By fusing multi-view RGB inputs with VGGT, selecting next-best-views via a VLM, and enforcing metric consistency through robot kinematics, GraspView produces reliable geometry for grasp planning and execution. Experiments show consistent improvements over RGB--D and single-view baselines, especially under occlusions, close-range sensing, and with transparent objects. These results demonstrate the practicality of RGB-only pipelines and suggest promising directions toward more general and robust robotic manipulation.

\bibliographystyle{IEEEtran}
\bibliography{references,IEEEabrv}

\end{document}